%% file: main.tex
\documentclass[10pt,twocolumn,letterpaper]{article}

\usepackage{wacv}
\usepackage{times}
\usepackage{epsfig}
\usepackage{graphicx}
\usepackage{amsmath}
\usepackage{amssymb}
\usepackage{subfig}
\usepackage[table]{xcolor}
\usepackage{tabularx}
\usepackage{csvsimple}
\usepackage{booktabs}
\usepackage{color}
\usepackage{microtype}
\usepackage[export]{adjustbox}
\newcolumntype{Y}{>{\centering\arraybackslash}X}
\usepackage{pdfpages}

\usepackage[pdfa,pagebackref=true,breaklinks=true,letterpaper=true,colorlinks,bookmarks=false]{hyperref}

\wacvfinalcopy 

\def\wacvPaperID{294} 

\ifwacvfinal\fi
\begin{document}

\title{Real-time Progressive 3D Semantic Segmentation for Indoor Scenes}

\author{
  Quang-Hieu Pham$^1$ \hspace{0.2in}
  Binh-Son Hua$^2$ \hspace{0.2in}
  Duc Thanh Nguyen$^3$ \hspace{0.2in}
  Sai-Kit Yeung$^4$
  \\
  $^1$Singapore University of Technology and Design
  \\
  $^2$The University of Tokyo \hspace{0.5in}
  $^3$Deakin University
  \\
  $^4$Hong Kong University of Science and Technology
}

\maketitle
\ifwacvfinal\thispagestyle{empty}\fi

\begin{abstract}
  The widespread adoption of autonomous systems such as drones and assistant
  robots has created a need for real-time high-quality semantic scene
  segmentation. In this paper, we propose an efficient yet robust technique for
  on-the-fly dense reconstruction and semantic segmentation of 3D indoor
  scenes. To guarantee (near) real-time performance, our method is built atop an
  efficient super-voxel clustering method and a conditional random field with
  higher-order constraints from structural and object cues, enabling progressive
  dense semantic segmentation without any precomputation. We extensively
  evaluate our method on different indoor scenes including kitchens, offices,
  and bedrooms in the SceneNN and ScanNet datasets and show that our technique
  consistently produces state-of-the-art segmentation results in both
  qualitative and quantitative experiments.
\end{abstract}

\input{intro}
\input{related}
\input{method}
\input{eval}
\input{remark}

{\small
  \bibliographystyle{ieee}
  \bibliography{ref}
}

\end{document}

%% file: intro.tex
\vspace{-0.2in}
\section{Introduction}

Recent hardware advances in consumer-grade depth cameras have made high-quality
reconstruction of indoor scenes feasible. RGB-D images have been used to boost
the robustness of numerous scene understanding tasks in computer vision, such as
object recognition, object detection, and semantic segmentation. While scene
understanding using color or RGB-D images is a well explored
topic \cite{silberman-nyud-eccv12, gupta-perceptual-cvpr13, long-fcn-cvpr15},
good solutions for the same task in the 3D domain have been highly sought after,
particularly, those can produce accurate and high-quality semantic segmentation.

\begin{figure}[t]
  \centering
  \includegraphics[width=0.75\linewidth]{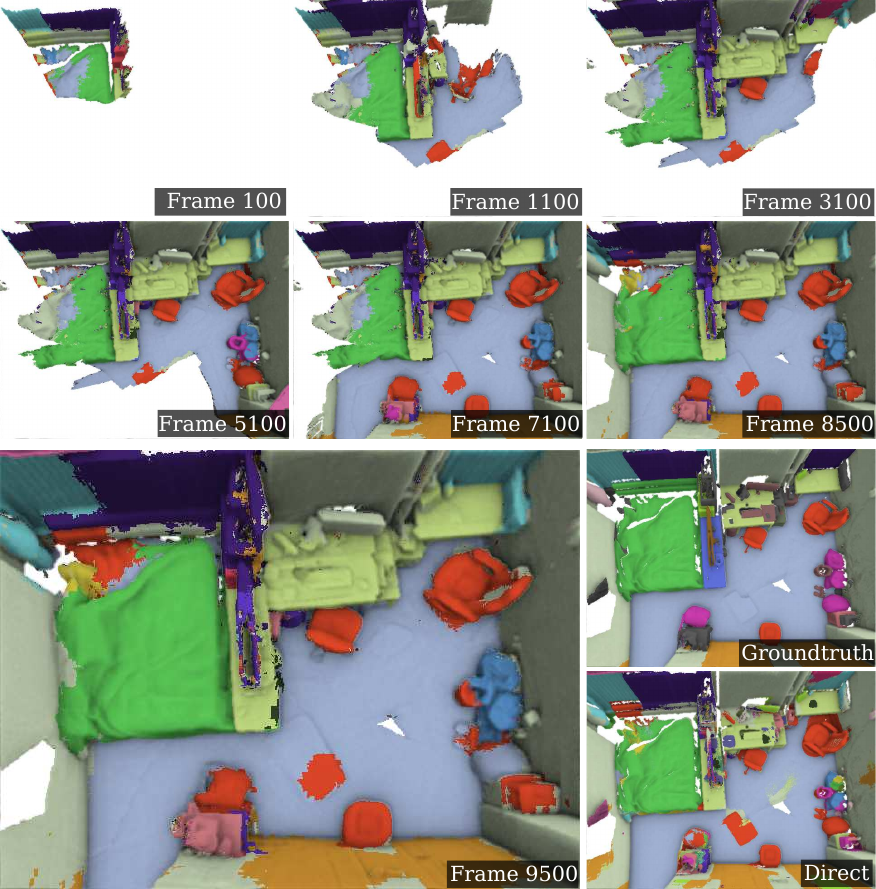}
  \caption{Progressive semantic segmentation of a 10K-frame bedroom scene in
    real time. Our method can resolve errors in segmentation while
    scanning. Note the segmentation error on the bed being gradually fixed as
    the user scans the scene.}
  \label{fig:teaser}
  \vspace{-0.2in}
\end{figure}

In this work, we propose a (near) real-time method for high-quality dense
semantic segmentation of 3D indoor scene. The backbone of our work is a
higher-order conditional random field (CRF) designed to infer optimal
segmentation labels from the predictions of a deep neural network. The CRF runs
in tandem with a revised pipeline for real-time 3D reconstruction using RGB-D
images as input. In contrast to traditional dense model, our CRF accepts
additional higher-order constraints from unsupervised object analysis, resulting
in high-quality segmentation. An example output from our proposed method is
shown in Figure \ref{fig:teaser}. Experiments proved that our method is capable
of producing high-quality semantic segmentation and achieve adequate temporal
consistency. In summary, our contributions are:
\begin{itemize}
\item A higher-order conditional random field that can resolve noisy predictions
  from a deep neural network into a coherent 3D dense segmentation, using
  additional object-level information.
\item An extended reconstruction pipeline, including an efficient voxel
  clustering technique, for efficient (near) real-time full-scene inference
  while scanning.
\item A thorough evaluation of state-of-the-art real-time semantic segmentation
  algorithms on two large scale indoor datasets, namely SceneNN
  \cite{hua-scenenn-3dv16} and ScanNet \cite{dai-scannet-cvpr17}.
\item Beyond category-based semantic segmentation, we also extend our method to
  instance-based semantic segmentation, and provide the first evaluation of
  real-time instance segmentation on SceneNN dataset.
\end{itemize}

%% file: related.tex
\section{Related Work}

\paragraph*{Indoor semantic segmentation.}
In their seminal work, Silberman \etal \cite{silberman-nyud-eccv12} proposed a
technique to segment cluttered indoor scenes into floor, walls, objects and
their support relationships. Their well-known NYUv2 dataset has since sparked
new research interests in semantic segmentation using RGB-D images.
Long \etal \cite{long-fcn-cvpr15} adapted neural networks originally trained for
classification to solve semantic segmentation by appending a fully connected
layer to the existing architecture. This method, however, tends to produce
inaccuracies along object boundaries. Since then, different techniques
\cite{yu-dilated-iclr16, badrinarayanan-segnet-pami17} has been proposed to
address this issue.
Some recent works also explored instance segmentation \cite{he-maskrcnn-iccv17,
  bai-watershed-cvpr17}, but such techniques only work in 2D.

In the 3D domain, a few datasets for 3D scene segmentation have also been
proposed \cite{hua-scenenn-3dv16, dai-scannet-cvpr17, armeni-s3dis-cvpr16}.
Early techniques focused on solving the problem by exploiting 3D volumes. For
example, Song \etal \cite{song-sscnet-cvpr17} and Dai \etal
\cite{dai-scancomplete-cvpr18} proposed a network architecture for semantic
scene segmentation and completion at the same time.
Point-based deep learning \cite{qi-pointnet-cvpr17, li-pointcnn-nips18,
  hua-pointwise-cvpr18, wang-dgcnn-arxiv18, huang-rsnet-cvpr18} took another
direction and attempted to learn point representation for segmentation directly
from unordered point clouds. While the results from these neural networks are
impressive, they only take as input a small point cloud of a few thousand
points. To address large-scale or structural point cloud, clustering techniques
such as super-points \cite{landrieu-superpoint-cvpr18} or hierarchical data
structures such as octree \cite{riegler-octnet-cvpr17} and kd-tree
\cite{klokov-escape-iccv17} have been proposed. Hybrid methods such as SEGCloud
\cite{tchapmi-segcloud-3dv17} turns the point clouds into volumes for prediction
with a neural network and then propagates the results back to the original point
cloud.

Instead of directly processing in 3D, multiple view techniques
\cite{kundu-monocular-eccv14, lawin-projective-caip17, ma-multiview-iros17,
  qi-3dgraph-iccv17, dai-3dmv-eccv18} focused on transferring 2D segmentation to
3D. Other methods further exploit object cues such as spatial context
\cite{engelmann-spatial-iccvw17}.
Our method is based on multi-view segmentation as such techniques scale better to
large-scale scenes. Concurrently, we also aim to achieve real-time performance
with progressive scene reconstruction.  We would focus our discussion to the
most relevant interactive and real-time techniques.

\vspace{-0.1in}
\paragraph*{Real-time semantic segmentation.}
Our real-time semantic segmentation system requires an online dense 3D
reconstruction system. KinectFusion \cite{newcombe-kinect-ismar11} showed us how
to construct such system. To overcome the spatial constraints in the original
KinectFusion implementation, which prohibits large-scale 3D scanning,
Nie{\ss}ner \etal \cite{niessner-voxhash-tog13} used voxel hashing to reduce the
memory footprint.
Valentin \etal \cite{valentin-mesh-cvpr13} proposed an interactive scanning
system where the segmentation is learnt from user inputs. Unlike them, our
method is completely automatic without the need of user interaction, and thus
more suitable for robotics applications.
Our method is based on a segmentation prediction with 2D deep neural networks, a
2D-3D label transfer and optimization with a conditional random field (CRF). To
our knowledge, the closest works to ours in this aspect is from the robotics
community \cite{hermans-dense-icra14, wolf-fast-icra15,
  vineet-incremental-icra15, mccormac-semanticfusion-icra17,
  held-probabilistic-rss16, yang-occupancy-iros17}. Early methods
\cite{hermans-dense-icra14, wolf-fast-icra15, vineet-incremental-icra15}
utilized random forest classifiers to initialize the CRF but their end-to-end
pipeline performance was far from real time. Similar to our approach, McCormac
\etal \cite{mccormac-semanticfusion-icra17} utilized segmentation predictions
from a deep neural network and achieved real-time performance on sparse point
cloud. In comparison, our method preserves surface information completely by
working with an on-the-fly sparse volume representation from Voxel Hashing
\cite{niessner-voxhash-tog13}, and introduce a higher-order conditional random
field model to refine 3D segmentation.

\vspace{-0.1in}
\paragraph*{Conditional random field.}
The CRF model, often containing unary and pairwise terms, is commonly used as
post-processing step \cite{chen-deeplab-pami18} to address noise in semantic
segmentation. Kr{\"a}henb{\"u}hl and Koltun \cite{krahenbuhl-densecrf-nips11}
proposed an efficient message passing method to perform inference on a
fully-connected model.
Recently, with the immense advances in deep learning, it is possible to embed
CRF into neural networks \cite{zheng-crfasrnn-iccv15, arnab-hocrf-eccv16} and
its parameters can be learnt jointly with the network via back-propagation.
While representing CRF by a recurrent neural network
\cite{zheng-crfasrnn-iccv15, arnab-hocrf-eccv16} is advantageous, applying such
end-to-end framework to our problem poses some challenges. First in the context
of progressive 3D reconstruction and segmentation, 2D predictions from multiple
views have to be combined to produce the labeling of 3D model, which is not
supported in the previous method where only the segmentation of one single image
is predicted. Second, their methods is computationally demanding which does not
fit our real-time requirement. Third, the number of 2D images used to calculate
the unaries is not fixed, compared to using only one input image as in previous
approaches. In this work, we instead run the CRF separately on 3D after
processing 2D semantic predictions from a convolutional neural network.

CRF is also extended with high-order potentials to further improve coherency in
the label prediction. For example, Zhu \etal \cite{zhu-cosegmentation-wacv14}
explored high-order CRF for co-segmentation on images.  Yang \etal
\cite{yang-occupancy-iros17} uses a hierarchical CRF with potentials from
super-pixels on images for fast outdoor scene segmentation.
The CRF model we propose in this work is a higher-order CRF that includes object
cues for indoor scenes and works in tandem with the geometry reconstruction.
Our idea is that to obtain a coherent, high-quality segmentation, vertices in
the same object should be consider as a whole in the model. Moreover, noises and
inconsistencies should be fixed regularly as the user scans through the scene.

%% file: method.tex
\begin{figure*}[t!]
  \begin{minipage}{\linewidth}
    \center
    \includegraphics[width=0.8\linewidth]{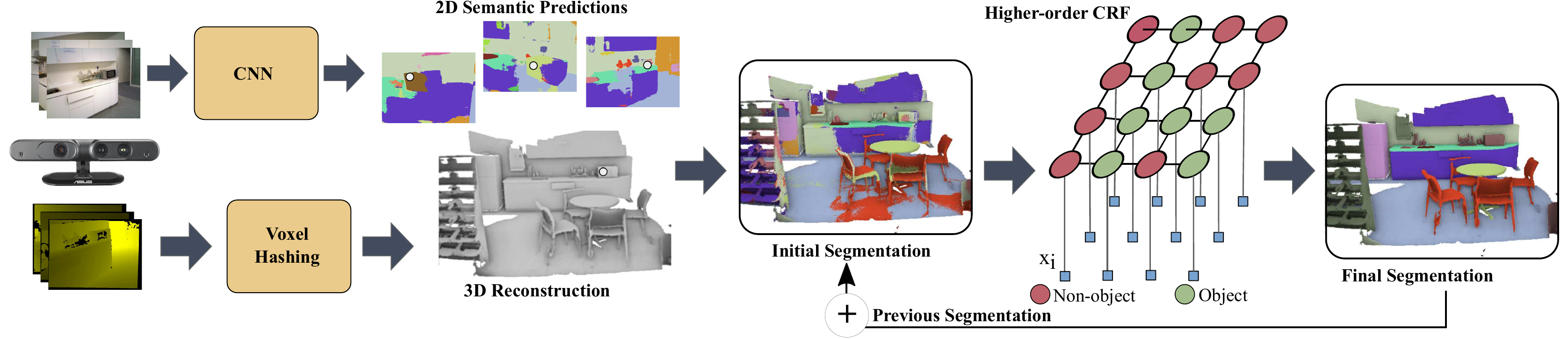}
  \end{minipage}
  \caption{Overview of our progressive indoor scene segmentation method. From
    continuous frames of an RGB-D sensor, our system performs on-the-fly
    reconstruction and semantic segmentation. All of our processing is performed
    on a frame-by-frame basis in an online fashion, thus useful for real-time
    applications.}
  \label{fig:overview}
  \vspace{-0.2in}
\end{figure*}

\section{Real-time RGB-D Reconstruction}
We now introduce our proposed method for the progressive dense semantic
segmentation problem. An overview of our framework is shown in
Figure~\ref{fig:overview}.

\subsection{Semantic label fusion}
Our online scanning system is built on top of the Voxel Hashing
\cite{niessner-voxhash-tog13} pipeline, reconstructing both geometric and
semantic information of the scene in real time. In principle, given an incoming
frame prediction from CNN, we must update the semantic label for each active
voxel accordingly, using the same integration process as described in
KinectFusion \cite{newcombe-kinect-ismar11}. For this problem, McCormac \etal
\cite{mccormac-semanticfusion-icra17} store a full discrete probability
distribution in each voxel, and update it by using recursive Bayesian
rule. However, doing so requires a large amount of memory and does not scale
well with large number of semantic classes. We employ the update process
proposed by Cavallari and Di Stefano \cite{cavallari-semanticfusion-eccvw16},
where each voxel only stores the current best label and its confidence.

\subsection{Progressive super-voxel clustering}
Now we explain in details our super-voxel clustering method, which will provide
a new domain to define our CRF with higher-order constraints.  Our super-voxel
clustering method resembles previous local k-means clustering techniques such as
VCCS \cite{papon-vccs-cvpr13} or SLIC \cite{achanta-slic-pami12}. The main
difference in our super-voxel clustering method is that, to amortize the
computation cost, we create super-voxels in a progressive manner, performing one
clustering iteration at a time, which will adapt better to the changes in the
current reconstructed scene. In our system, we consider common features such as
voxel color and position to define the distance measure $D$:
\begin{align}
  \label{eq:distance}
  D = \sqrt{\frac{\alpha D_c}{n_c} + \frac{\beta D_s}{n_s}}
\end{align}
where $D_c$ and $D_s$ are the color and spatial distances, with $n_c$ and $n_s$
act as the normalizers; $\alpha$ and $\beta$ control the relative weighting of
color and spatial distances. In all of our experiments, we set $\alpha$ and
$\beta$ to 1; the normalization values $n_c$ and $n_s$ are based on the chosen
voxel size which is $0.008m$ and the CIELab color space. Here one can further
utilize voxel normals for the distance measure but we found that the quality of
the clustering does not improve much despite of the expensive cost to compute
normals per voxel.
Another possible extension is to consider features provided by the 2D semantic
segmentation network in the distance measure. However, the memory storage per
voxel would be very costly because each feature vector often has at least tens
of floating point numbers. Some compressions might help in this case.

Suppose that an existing set of super-voxels are already provided. For an
incoming RGB-D frame at time $t$, after camera pose estimation, we can find out
the current active set $V_t$ of voxels using an inside/outside check on the
current camera frustum. Our goal is to assign each of these voxels into a
super-voxel (or cluster). This process is as follows: first new seeds are
sampled on uninitialized regions, based on a chosen spatial interval $S$. For
each active voxel, we assign it to the nearest cluster according to the distance
in Equation \ref{eq:distance}. Next, we update the centers information based on
the new cluster assignment. This process is repeated for every incoming RGB-D
frame, providing a ``live'' unsupervised over-segmentation of the scene.

Our progressive super-voxel building scheme fits well into the common dense
RGB-D reconstruction pipelines such as KinectFusion
\cite{newcombe-kinect-ismar11} or Voxel Hashing \cite{niessner-voxhash-tog13},
and can be implemented efficiently on the GPU. In practice, we only consider
voxels close to the surface, based on their distance-to-surface
values. Performing inference on these super-voxels significantly reduces the
domain size of our CRF, and thus paves the way for real-time semantic
segmentation.

\subsection{Real-time object proposal}
For 3D object proposal, Karpathy \etal \cite{karpathy-objectness-icra13}
presented a method for discovering object models from 3D meshes of indoor
environments. Their method first generates object candidates by over-segmenting
the scene on different thresholds. The candidates are then evaluated and
suppressed based on geometric metrics to produce the final proposals. Kanezaki
\cite{kanezaki-ss3d-iros15} proposed an extension of selective search for object
proposal on 3D point cloud.

One common drawback of these methods is their high computation cost, since they
require a costly object analysis on different scales. This process has to be
done for every update, which hinders real-time performance. In this work, we
explore on a new direction for object proposal, in which we propose object based
on statistical evidences.

Our object proposal is come from a simple observation: given an object and
multiple observations, it should be identified as an object in most of the
corresponding 2D semantic predictions. Hence, for each incoming RGB-D frame, we
update the objectness score of a voxel given its current predicted label.
Specifically, we decrease the objectness score if the prediction is a non-object
label,~\ie~wall, floor, or ceiling; and increase it otherwise. To perform object
proposal, we employ an efficient graph-based segmentation algorithm from
Felzenszwalb and Huttenlocher \cite{felzenszwalb-graph-ijcv04}.  The edge
weight between two super-voxels $i$ and $j$ is defined as
$w_{i,j}=w^{\alpha}_{i,j}+w^{\eta}_{i,j}+w^{\omega}_{i,j}$ where
$w^{\alpha}_{i,j}$, $w^{\eta}_{i,j}$, and $w^{\omega}_{i,j}$ are the edge weight
for voxel color, normal, and objectness, respectively. We normalize the each of
the weights accordingly. To reduce computation cost, we only compute the terms
using representative values from super-voxel centroids.

\section{Higher-order CRF Refinement}
\vspace{-0.1in}
Using CRF as a post-processing step is a common technique in semantic
segmentation. However, for real-time applications, there are two limitations
that we must address. First is the classification errors caused by
inconsistencies, sometimes known as ``bleeding'', that is also reported by
Valentin \etal \cite{valentin-semanticpaint-tog15}. The second issue is
scalability, since the number of vertices in the graph grows to millions during
scanning, causing CRF optimizations to become much slower over time. In this
work, we address both limitations by introducing a CRF model with
\emph{higher-order constraints} on \emph{super-voxels} to perform online
segmentation. This model is lightweight and very easy to compute, allowing it to
work on a wide range of indoor scenes, while remaining computationally efficient
for real-time use.

Let $\mathcal{M}^t$ be the 3D geometry at time $t$ with $N^t$ super-voxels. In a
semantic segmentation problem, we attempt to assign every super-voxel with a
label from a discrete label space, denoted $\mathcal{L}=\{l_1,l_2, \dots
,l_L\}$. Let $\mathbf{X}^t = \{x^t_1, \dots, x^t_N\}$ define a set of random
variables, one for each super-voxel, where $x^t_i \in \mathcal{L}$. An assignment
of every $x^t_i$ will be a solution to the segmentation problem at time $t$. For
shorter notation, we will drop the superscript time notation from now on.

Given the above definitions, we define a graph $\mathcal{G}$ where each vertex
is from $\mathbf{X}$. In addition, let $\mathcal{C}$ be the set of cliques in
$\mathcal{G}$, given by an object proposal method. For every clique $r \in
\mathcal{C}$, we can select a corresponding set of random variables
$\mathbf{x}_r$ that belongs to $r$. Our CRF model introduces three new types of
higher-order potential, namely objectness potential $\psi^{O}$, consistency
potential $\psi^{C}$ and object relationship potential $\psi^{R}$. These terms
are later explained in Section \ref{sec:objectness}, \ref{sec:consistency}, and
\ref{sec:relationship}, respectively. Our complete CRF model is then defined as
\begin{equation}
  \label{eq:energy_function}
  \begin{split}
    E(\mathbf{X}) &=
    \sum_{i}{\varphi(x_i)} + \sum_{i<j}{\psi^{P}(x_i,x_j)} \\
    &+ \sum_{r \in \mathcal{C}}{\psi^{O}(\mathbf{x}_r)}
    + \sum_{r \in \mathcal{C}}{\psi^{C}(\mathbf{x}_r)} 
    + \sum_{r,q \in \mathcal{E}(\mathcal{C})}{\psi^{R}(\mathbf{x}_r,\mathbf{x}_q)}
  \end{split}
\end{equation}
where $\varphi(x_i)$ and $\psi^{P}(x_i,x_j)$ are the unary and pairwise terms
used in the traditional dense CRF model. The unary term represent the prediction
from a local classifier. In our case, it is obtained from fusing CNN predictions
during reconstruction.

The pairwise (smoothness) potential $\psi^{P}(x_i,x_j)$ is parameterized by a
Gaussian kernel
\begin{align}
  \psi^{P}(x_i, x_j) &= \mu_{ij}\exp\left( - \frac{ \mid p_i - p_j \mid }{ 2\theta^2_\alpha }
  - \frac{ \mid n_i - n_j \mid }{ 2\theta^2_\beta } \right)
\end{align}
where $\mu_{ij}$ is the label compatibility function between $x_i$ and $x_j$
given by the Potts model; $p_i$ and $n_i$ are the location and normal of the
$i^{th}$ super-voxel; $\theta_\alpha$ and $\theta_\beta$ are standard deviations
of the kernel.

\subsection{Objectness potential}
\label{sec:objectness}
The term $\psi^{O}(\mathbf{x}_r)$ captures the mutual agreement between the
objectness score of a clique and its semantic label. Ideally, we would want a
clique with low objectness score to take a non-object label,~\ie~wall, floor, or
ceiling; and inversely. To model the objectness potential of a clique, we first
introduce latent binary random variables $y_1, \dots, y_{\mid \mathcal{C}\mid}$.
$y_k$ can be interpreted as follows: if the $k^{th}$ proposal has been found to
be an object, then $y_k$ is 1, otherwise it will be 0. Let $\mathcal{O}$ be the
subset of $\mathcal{L}$, which comprises of object classes in the label space.
We can then define our objectness potential
\begin{align}
  \psi^{O}(\mathbf{x}_r) =
  \begin{cases}
    \frac{1}{\mid \mathbf{x}_r \mid}\sum_{i \in \mathbf{x}_r}{[x_i \notin \mathcal{O}]}, & \text{if } y_r = 1, \\
    \frac{1}{\mid \mathbf{x}_r \mid}\sum_{i \in \mathbf{x}_r}{[x_i \in \mathcal{O}]}, & \text{if } y_r = 0, \\
  \end{cases}
\end{align}
where $[ \cdot ]$ is a function that converts a logical proposition into $1$ if
the condition is satisfied, otherwise it would be $0$. The purpose of this term
is to correct misclassification errors in the local classifier, based on
external unsupervised information from object proposal.

\subsection{Label consistency}
\label{sec:consistency}
The term $\psi^{C}(\mathbf{x}_r)$ enforces regional consistency in semantic
segmentation. Since we want vertex labels in the same clique to be homogeneous,
the cost function penalizes label based on its frequency in the clique. Let
$f_r(l_k)$ be the normalized frequency of label $l_k \in \mathcal{L}$ inside the
$r^{th}$ clique, which is of the range between $0$ and $1$. The consistency cost
will be the entropy of the underlying distribution:
\begin{align}
  \psi^{C}(\mathbf{x}_r) = - \sum_{l_k \in \mathcal{L}}{f_r(l_k)\log{f_r(l_k)}}
  \label{eq:label_consistency}
\end{align}
This term dampens infrequent labels in a clique. In experiments, We observed
that the label consistency cost helps fixing low frequency errors in the output
segmentation.

\subsection{Region relationship}
\label{sec:relationship}
The relationship potential $\psi^{R}$ encodes the relation between two regions
(cliques) and their semantic labels. This cost is applied on neighboring
regions, based on super-voxel connectivity. In our model, the term
$\psi^{R}(\mathbf{x}_r,\mathbf{x}_q)$ is defined based on the co-occurrence of
class labels in the regions. Specifically, let $\mathcal{E}(\mathcal{C}) \subset
\mathcal{C} \times \mathcal{C}$ be the edges between connected cliques. The
object relationship cost between $\mathbf{x}_r$ and $\mathbf{x}_q$ is defined as
follows,
\begin{align}
\label{eq:region_to_region}
    &\psi^{R}(\mathbf{x}_r, \mathbf{x}_q) = -\sum_{l_i \in \mathcal{L}} \sum_{l_j \in \mathcal{L}}
    \log \left( f_r(l_i) f_q(l_j) \Lambda_{l_i,l_j} \right)
\end{align}
where $\Lambda_{l_i,l_j}$ is the co-occurrence cost based on the class labels
$l_i$ and $l_j$ and designed such that the more often $l_i$ and $l_j$ co-occur,
the greater $\Lambda_{l_i,l_j}$ is. This cost acts like a prior to prevent
uncommon label transition,~\eg~chair to ceiling, ceiling to floor, etc; and can
be learnt beforehand. $f_r$ and $f_q$ are the label frequencies, as presented in
(\ref{eq:label_consistency}).

In our CRF model, each term is accompanied with a weight to balance their values
that we omit them in our formulas for better clarity. We learn these weights by
grid search, and keep them unchanged in all of the experiments.

Finally, semantic segmentation can be done by minimizing the energy function
$E(\mathbf{X})$ defined in (\ref{eq:energy_function}). In this paper, we adopt
the variational mean field method \cite{krahenbuhl-densecrf-nips11} for
efficiently optimizing $E(\mathbf{X})$. Details of the inference process can be
found in the supplementary material.

\subsection{Temporal consistency}
We support temporal consistency with a simple modification of the unary term as
follows. To minimize storage, let us only consider time $t-1$ and time $t$. The
unary term becomes a weighted sum that takes as input the final labels at time
$t - 1$ ($\mathbf{X}_{CRF}$, after CRF of time $t-1$) and the CNN predicted
labels at the time $t$ ($\mathbf{X}_{predicted}$, before CRF):
$\mathbf{X}_{unary}^t = \tau \mathbf{X}_{predicted}^t +
(1-\tau)\mathbf{X}_{CRF}^{t-1}$ where $\mathbf{X}$ are the label probabilities,
and $\tau \in [0, 1]$ is a scalar value. Smaller $\tau$ favors temporal
consistency. We set $\tau$ empirically by plotting the segmentation accuracy
with multiple $\tau$. Our experiment (see supplementary) shows that $\tau=0.5$
strikes a balance between accuracy and temporal consistency.

\subsection{Instance segmentation}
Beyond category-based semantic segmentation, we extend our technique to support
instance-based semantic segmentation in real time, which we refer to as
\emph{instance segmentation} for brevity. The key change is that CRF model now
outputs instance IDs instead of class segmentation labels. Other terms and the
optimization process are kept unchanged.

A straightforward approach for instance segmentation would be utilizing a deep
neural network that can perform instance-based segmentation in 2D, and then
propagate the predictions from 2D to 3D as in the category-based semantic
segmentation case. However, this approach requires us to track the instance IDs
over time, which is in fact a challenging problem, since the networks, e.g.,
\cite{he-maskrcnn-iccv17}, can only predict one frame at a time.

Our solution is to combine category-based semantic segmentation network with the
following instance-based segmentation to yield instance IDs. For each vertex
$x_i$ in the CRF, we have to define probabilities over every possible instance
IDs. The label space, $\mathcal{L}=\{l_1,l_2, \dots ,l_L\}$, would be the set of
all instance IDs in the current 3D reconstruction. Performing CRF inference on
the entire set of instance IDs would be infeasible. Here we reduce the problem
size by first filtering out the instance IDs that are not in the current camera
frustum at time $t$, giving a reduced label space $\mathcal{L}^t$. Our
higher-order CRF will only optimize instance labels of super-voxels in the camera
frustum, instead of the entire scene as before. The result is then fused into
the current model.

Another issue in progressive instance segmentation is how to update the label
space $\mathcal{L}$, since online scanning will continuously introduce new
instances to our model. We tackle this problem by creating a special
\emph{unknown} instance ID. All of the newly scanned voxels will be initialized
with unknown. After each CRF inference step, the largest connected component,
which is based on category, belongs to the unknown instance will be spawned as a
new instance. We also update the set of instance IDs accordingly.

%% file: eval.tex
\section{Experiments}
\begin{table}[t]
  \scriptsize
  \centering
  \begin{filecontents*}{figures/csv/online.csv}
Scene,DirectA,DirectIoU,SemanticFusionA,SemanticFusionIoU,OursA,OursIoU,OursInstanceA
011,0.770,0.654,0.776,0.664,\textbf{0.800},\textbf{0.699},0.521
016,0.607,0.476,0.625,0.497,\textbf{0.680},\textbf{0.566},0.342
030,0.584,0.450,0.597,0.466,\textbf{0.658},\textbf{0.559},0.568
061,0.751,0.606,0.777,0.641,\textbf{0.809},\textbf{0.691},0.591
078,0.497,0.365,0.515,0.388,\textbf{0.535},\textbf{0.454},0.349
086,0.622,0.519,0.646,0.551,\textbf{0.668},\textbf{0.587},0.350
096,0.659,0.536,\textbf{0.668},0.549,0.666,\textbf{0.565},0.265
206,0.766,0.654,\textbf{0.778},0.673,0.775,\textbf{0.675},0.417
223,0.669,0.559,0.689,0.587,\textbf{0.729},\textbf{0.655},0.409
255,0.423,0.316,0.439,0.334,\textbf{0.558},\textbf{0.473},0.486
\end{filecontents*}
\csvreader[
    head to column names,
    tabular=r  cc  cc  cc,
    table head=
    \toprule
    ID &
    \textbf{Direct} &
    \textbf{SF} &
    \multicolumn{2}{c}{\textbf{Ours}} \\
    & Class & Class & Class & Instance
    \\\midrule,
    table foot=\bottomrule]{figures/csv/online.csv}
    { Scene=\Scene,
      DirectA=\DirectA,
      SemanticFusionA=\SemanticFusionA,
      OursA=\OursA,
      OursInstanceA=\OursInstanceA,
    }
    {
      \Scene &
      \DirectA &
      \SemanticFusionA &
      \OursA &
      \OursInstanceA
    }
    \caption{Comparison of category-based semantic segmentation accuracy on
      typical scenes in SceneNN dataset. We report performances on office,
      kitchen, bedroom, and other scenes. Our proposed CRF model consistently
      outperforms the naive approach that directly fuses neural network
      predictions to 3D (Direct)~\cite{cavallari-semanticfusion-eccvw16}, and
      SemanticFusion (SF)~\cite{mccormac-semanticfusion-icra17}. Please also
      refer to the supplementary document for weighted IoU scores. The final
      column reports the average precision scores of our instance-based
      segmentation results.}
    \label{tab:online}
    \vspace{-0.2in}
\end{table}

\begin{table}[t]
  \scriptsize
  \centering
  \begin{filecontents*}{figures/csv/offline.csv}
Scene,DirectSegNetA,DirectSegNetIoU,OursSegNetA,OursSegNetIoU,DirectFCNA,DirectFCNIoU,OursFCNA,OursFCNIoU,DirectSSCNetA,DirectSSCNetIoU,OursSSCNetA,OursSSCNetIoU
011,0.747,0.622,\textbf{0.837},\textbf{0.783},0.667,0.550,0.743,0.682,0.475,0.299,0.497,0.378
016,0.556,0.425,0.714,0.636,0.580,0.463,0.623,0.519,0.648,0.514,\textbf{0.798},\textbf{0.679}
030,0.554,0.413,0.668,0.570,0.584,0.421,\textbf{0.704},\textbf{0.624},0.505,0.418,0.510,0.383
061,0.549,0.351,\textbf{0.841},\textbf{0.728},0.324,0.162,0.457,0.276,0.700,0.640,0.693,0.641
078,0.542,0.405,\textbf{0.666},\textbf{0.594},0.551,0.393,0.663,0.562,0.515,0.374,0.588,0.428
086,0.587,0.459,\textbf{0.686},\textbf{0.613},0.491,0.356,0.631,0.551,0.543,0.464,0.517,0.448
096,0.615,0.495,\textbf{0.683},\textbf{0.605},0.577,0.460,0.619,0.545,0.631,0.545,0.658,0.567
206,0.659,0.502,0.812,0.718,0.626,0.463,0.828,0.729,\textbf{0.861},\textbf{0.800},0.834,0.771
223,0.648,0.524,0.758,\textbf{0.685},0.693,0.567,\textbf{0.760},0.669,0.644,0.528,0.639,0.526
255,0.521,0.410,0.654,\textbf{0.559},0.577,0.429,\textbf{0.718},0.580,0.547,0.450,0.661,0.526
\end{filecontents*}
\csvreader[
    head to column names,
    tabular=r cc cc cc,
    table head=
    \toprule
    \textbf{Acc.} &
    \multicolumn{2}{c}{\textbf{SegNet}} &
    \multicolumn{2}{c}{\textbf{FCN-8s}} &
    \multicolumn{2}{c}{\textbf{SSCNet}} \\
    ID~ & Base & Ours & Base & Ours & Base & Ours\\
    \midrule,
    table foot=\bottomrule]{figures/csv/offline.csv}
    { Scene=\Scene,
      DirectSegNetA=\DirectSegNetA,
      DirectSegNetIoU=\DirectSegNetIoU,
      OursSegNetA=\OursSegNetA,
      OursSegNetIoU=\OursSegNetIoU,
      DirectFCNA=\DirectFCNA,
      DirectFCNIoU=\DirectFCNIoU,
      OursFCNA=\OursFCNA,
      OursFCNIoU=\OursFCNIoU,
      DirectSSCNetA=\DirectSSCNetA,
      DirectSSCNetIoU=\DirectSSCNetIoU,
      OursSSCNetA=\OursSSCNetA,
      OursSSCNetIoU=\OursSSCNetIoU
    }
    {
      \Scene &
      \DirectSegNetA & \OursSegNetA &
      \DirectFCNA & \OursFCNA &
      \DirectSSCNetA & \OursSSCNetA
    }
    \caption{Accuracy scores of offline semantic segmentation task on
      SceneNN~\cite{hua-scenenn-3dv16}. Our proposed CRF model consistently
      improves the accuracy of the initial predictions from SegNet,
      FCN-8s~\cite{long-fcn-cvpr15} and SSCNet~\cite{song-sscnet-cvpr17}.
      Please refer to the supplementary document for weighted IoU scores and more
      results on ScanNet~\cite{dai-scannet-cvpr17}.}
    \label{tab:offline}
    \vspace{-0.2in}
\end{table}

\paragraph{Experiment setup.}
In SemanticFusion~\cite{mccormac-semanticfusion-icra17}, the authors chose to
evaluate on NYUv2 dataset, a popular 2D dataset for semantic segmentation
task. However, evaluation in 2D by projecting labels from 3D model to 2D image
is not completely sound; since 2D images cannot cover the entire scene, and
there are potential ambiguities when doing 2D-3D projection. To tackle this
problem, we perform our evaluation on SceneNN~\cite{hua-scenenn-3dv16} and
ScanNet~\cite{dai-scannet-cvpr17}, which are two 3D mesh datasets with dense
annotations. Our evaluation can act as a reference benchmark for real-time 3D
scene segmentation systems.

We adopt two common metrics from 2D semantic segmentation for our 3D evaluation,
namely vertex accuracy (\emph{A}) and frequency weighted intersection over union
(\emph{wIoU}). Due to space constraint, we only show our accuracy evaluation in
this section. Please refer to our supplementary document for the wIoU evaluation.

\vspace{-0.1in}
\paragraph{Implementation details.}
To get the 2D segmentation predictions, we use SegNet
\cite{badrinarayanan-segnet-pami17} trained on SUN RGB-D dataset. We chose
SegNet as it has better accuracy for indoor scenes but more compact and faster
alternatives \cite{paszke-enet-arxiv16, wu-sparsity-arxiv17,
  li-difficulty-cvpr17, romera-convnet-iv17, zhao-pspnet-cvpr17,
  yu-bisenet-eccv18, zhao-icnet-eccv18} could be used. The CRF inference is the
work by Kr{\"a}henb{\"u}hl and Koltun \cite{krahenbuhl-densecrf-nips11}. For the
best performance and responsiveness for real time use, we run one iteration of
the CRF inference in each frame, and the CNN predictions every K frames (with K
= 10 in our experiment). This aligns with the fact that the geometry change is
usually subtle in each frame, and label propagation with CRF per frame is
sufficient for a good prediction while maximizing responsiveness. After K frames
when geometry changes more significantly, we update the segmentation with the
more accurate but costly CNN predictions.

\vspace{-0.1in}
\paragraph{Online semantic segmentation.}
We compare our approach to the following methods: (1) Direct label fusion
\cite{cavallari-semanticfusion-eccvw16}; and (2) SemanticFusion
\cite{mccormac-semanticfusion-icra17}. To give a fair comparison, all of our
online results are reconstructed using the same camera trajectories and semantic
predictions from SegNet.

We present the performance comparison of our algorithm in various indoor
settings. Results are shown in Table \ref{tab:online}. Our method outperforms
SemanticFusion and the direct fusion approach in almost all of the scenes.
Qualitative results also show that our method is less subjective to noise and
inconsistencies in segmentation compared to other approaches, especially on
object boundaries.

\vspace{-0.1in}
\paragraph{Offline semantic segmentation.}
We further investigate our model robustness subject to different types of
initial segmentation. We perform the experiment in offline setting, taking unary
predictions from different neural networks and refine them using our proposed
higher-order CRF. For the offline experiment, since the meshes are already
provided, we run CRF inference directly on a per-vertex level to produce highest
segmentation quality. All of the neural networks are trained on the NYUv2
dataset \cite{silberman-nyud-eccv12}.

\begin{figure*}[t!]
  \centering
  \includegraphics[width=0.9\linewidth]{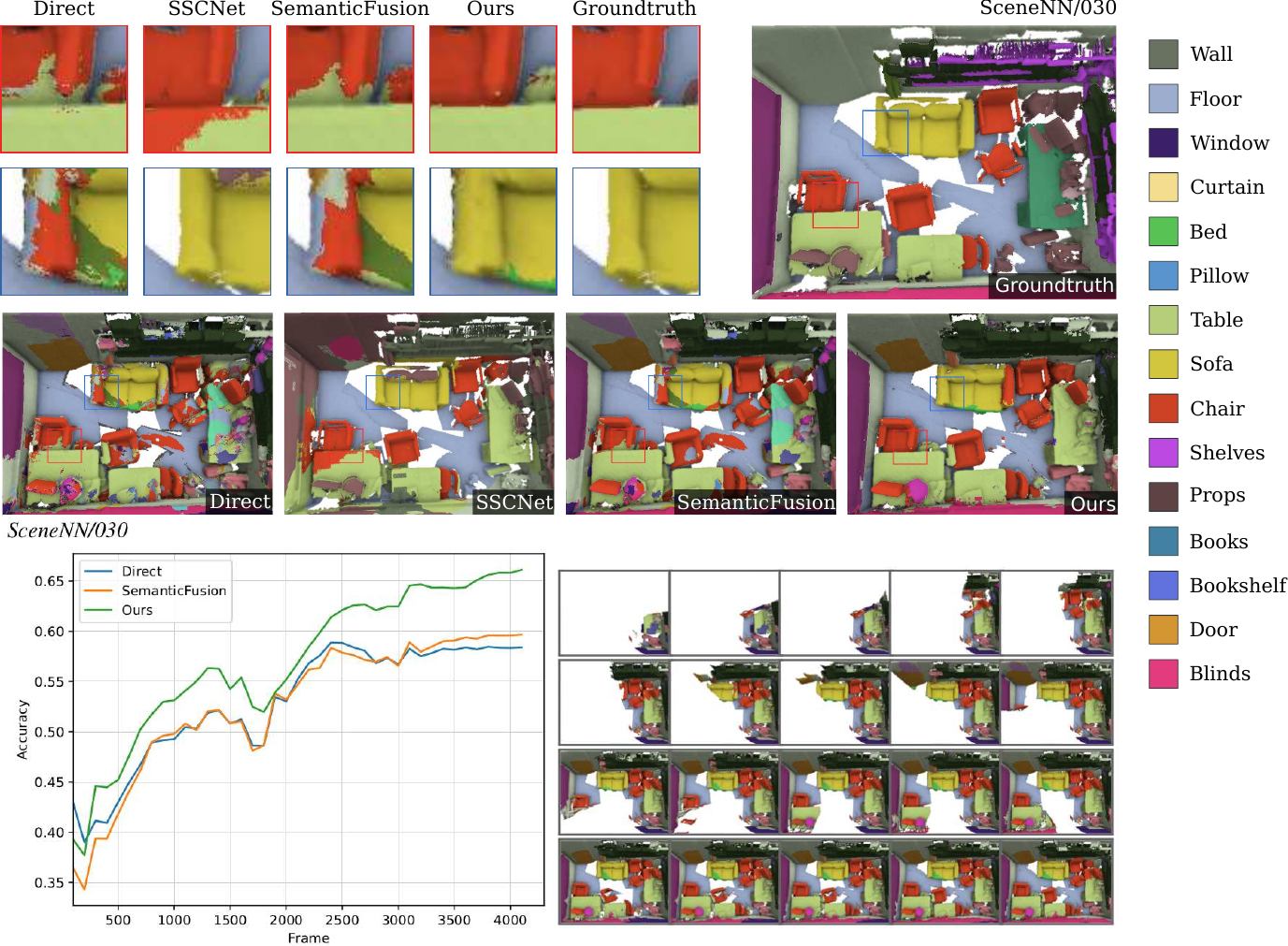}
  \caption{Qualitative results and a temporal accuracy on a selected scene from
    SceneNN. The top right image is the ground truth segmentation. The results
    from direct fusion \cite{cavallari-semanticfusion-eccvw16}, SSCNet
    \cite{song-sscnet-cvpr17}, SemanticFusion
    \cite{mccormac-semanticfusion-icra17} and ours are shown on the second row,
    respectively. The respective progressive semantic segmentation results of
    our method are shown on the bottom right. Please refer to the supplementary
    materials for the full qualitative results.}
  \label{fig:seg3d}
  \vspace{-0.2in}
\end{figure*}

Results from SegNet \cite{badrinarayanan-segnet-pami17}, SSCNet
\cite{song-sscnet-cvpr17}, and FCN-8s \cite{long-fcn-cvpr15} are shown in Table
\ref{tab:offline}. Note that SSCNet produces a $60\times36\times60$ volume low
resolution segmentation for entire scene due to memory constraints, so we need
to re-sample to a higher resolution. In contrast, our 2D-to-3D approach can
achieve segmentation on high resolution meshes at almost real-time rate. Again,
our method improves SegNet by 10\% in accuracy, SSCNet by 8\%, and FCN by 9\%.
This shows that our proposed CRF performs robustly to different kinds of unary.
See Figure \ref{fig:seg3d} for more detailed qualitative comparisons.

\vspace{-0.1in}
\paragraph{Per-class accuracy.}
We measured per-class accuracy of our method and SemanticFusion
\cite{mccormac-semanticfusion-icra17} (see Table \ref{tab:perclass} below). The
results show that our method consistently outperforms SemanticFusion. On
average, we increase accuracy by 6\% compared to SemanticFusion and 11\%
compared to the direct fusion method.

\definecolor{wall}{RGB}{198,213,181}
\definecolor{floor}{RGB}{164,181,215}
\definecolor{cabinet}{RGB}{97,57,195}
\definecolor{bed}{RGB}{90,200,86}
\definecolor{chair}{RGB}{224,71,40}
\definecolor{sofa}{RGB}{213,201,63}
\definecolor{table}{RGB}{102,143,128}
\definecolor{door}{RGB}{212,150,51}
\definecolor{window}{RGB}{67,35,119}
\definecolor{bookshelf}{RGB}{97,113,222}
\definecolor{picture}{RGB}{161,225,66}
\definecolor{counter}{RGB}{112,224,170}
\definecolor{blinds}{RGB}{227,60,126}
\definecolor{desk}{RGB}{74,152,112}
\definecolor{shelves}{RGB}{186,73,225}
\definecolor{curtain}{RGB}{122,218,223}
\definecolor{dresser}{RGB}{215,170,84}
\definecolor{pillow}{RGB}{89,148,207}
\definecolor{mirror}{RGB}{167,71,33}
\definecolor{floor-mat}{RGB}{180,108,207}
\definecolor{clothes}{RGB}{101,130,50}
\definecolor{ceiling}{RGB}{214,105,165}
\definecolor{books}{RGB}{66,129,164}
\definecolor{fridge}{RGB}{48,65,36}
\definecolor{television}{RGB}{137,99,42}
\definecolor{paper}{RGB}{96,92,156}
\definecolor{towel}{RGB}{225,131,87}
\definecolor{shower-curtain}{RGB}{49,32,65}
\definecolor{box}{RGB}{212,126,132}
\definecolor{whiteboard}{RGB}{130,53,111}
\definecolor{person}{RGB}{212,176,127}
\definecolor{night-stand}{RGB}{157,48,85}
\definecolor{toilet}{RGB}{94,153,161}
\definecolor{sink}{RGB}{125,44,38}
\definecolor{lamp}{RGB}{220,182,191}
\definecolor{bathtub}{RGB}{74,35,37}
\definecolor{bag}{RGB}{207,62,178}
\definecolor{structure}{RGB}{63,84,106}
\definecolor{furniture}{RGB}{110,123,97}
\definecolor{prop}{RGB}{140,99,101}

\begin{table*}[ht]
  \scriptsize
  \begin{tabularx}{\linewidth}{|l|Y|Y|Y|Y|Y|Y|Y|Y|}
    \hline
    & \cellcolor{wall}wall & \cellcolor{floor}floor & \cellcolor{cabinet}\textcolor{white}{cabinet}
    & \cellcolor{bed}bed & \cellcolor{chair}chair & \cellcolor{sofa}sofa
    & \cellcolor{table}table & \cellcolor{door}door \\
    \hline
    Direct & 0.710 & 0.914 & 0.471 & 0.309 & 0.430 & 0.555 & 0.557 & 0.313 \\
    SF & 0.728 & 0.944 & 0.570 & 0.343 & 0.463 & 0.578 & \textbf{0.701} & 0.386 \\
    Ours & \textbf{0.750} & \textbf{0.965} & \textbf{0.620} & \textbf{0.375} & \textbf{0.649} & \textbf{0.661} & 0.698 & \textbf{0.513} \\
    \hline
    & \cellcolor{window}\textcolor{white}{window} & \cellcolor{bookshelf}\textcolor{white}{bookshelf}
    & \cellcolor{picture}picture & \cellcolor{counter}counter & \cellcolor{blinds}blinds
    & \cellcolor{desk}desk & \cellcolor{curtain}curtain & \cellcolor{pillow}pillow\\
    \hline
    Direct & 0.252 & 0.839 & 0.202 & 0.266 & 0.215 & 0.236 & 0.643  & 0.268 \\
    SF & 0.315 & 0.940 & \textbf{0.225} & 0.371 & 0.210 & 0.281 & 0.830 & \textbf{0.294} \\
    Ours & \textbf{0.425} & \textbf{0.947} & 0.121 & \textbf{0.551} & \textbf{0.231} & \textbf{0.408} & \textbf{1.000} & 0.253 \\
    \hline
    & \cellcolor{clothes}clothes & \cellcolor{ceiling}ceiling
    & \cellcolor{books}books & \cellcolor{fridge}\textcolor{white}{fridge}
    & \cellcolor{television}\textcolor{white}{television} & \cellcolor{paper}\textcolor{white}{paper}
    & \cellcolor{night-stand}\textcolor{white}{nightstand} & \cellcolor{sink}\textcolor{white}{sink} \\
    \hline
    Direct & 0.197 & 0.705 & 0.426 & 0.700 & 0.212 & 0.119 & 0.076 & 0.380 \\
    SF & 0.236 & 0.800 & 0.524 & 0.803 & 0.277 & \textbf{0.320} & 0.090 & \textbf{0.388} \\
    Ours & \textbf{0.290} & \textbf{0.858} & \textbf{0.603} & \textbf{0.823} & \textbf{0.643} & 0.097 & \textbf{0.145} & 0.342 \\
    \hline
    & \cellcolor{lamp}lamp & \cellcolor{shelves}\textcolor{white}{shelves}  & \cellcolor{bag}\textcolor{white}{bag}
    & \cellcolor{structure}\textcolor{white}{structure} & \cellcolor{furniture}\textcolor{white}{furniture} & \cellcolor{prop}\textcolor{white}{prop} & Average & \\
    \hline
    Direct & 0.284 & 0.000 & 0.226 & 0.121 & \textbf{0.110} & 0.275 & 0.367 & \\
    SF & 0.391 & 0.000 & 0.214 & 0.169 & 0.016 & 0.291 & 0.423 & \\
    Ours & \textbf{0.583} & 0.000 & \textbf{0.364} & \textbf{0.262} & 0.018 & \textbf{0.312} & \textbf{0.484} & \\
    \hline
  \end{tabularx}
  \caption{Per-class accuracy of 40 NYUDv2 classes on SceneNN dataset from
    direct fusion, SemanticFusion (SF) and ours. Note that some of the classes
    are missing from the evaluation data. Best view in color.}
  \label{tab:perclass}
\end{table*}

\begin{figure*}[t!]
  \centering
  \def\hsp{0.005in}
  \begin{minipage}{0.23\linewidth}
    \includegraphics[width=\linewidth]{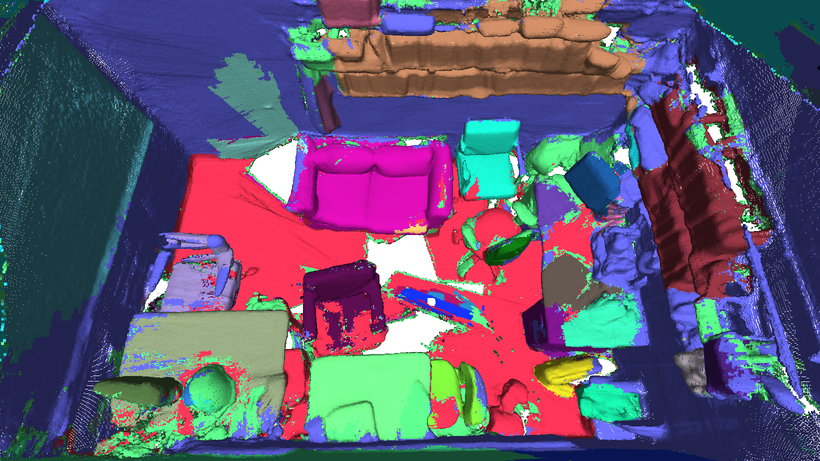}
    \centerline{Prediction}
  \end{minipage}%
  \begin{minipage}{0.23\linewidth}
    \includegraphics[width=\linewidth]{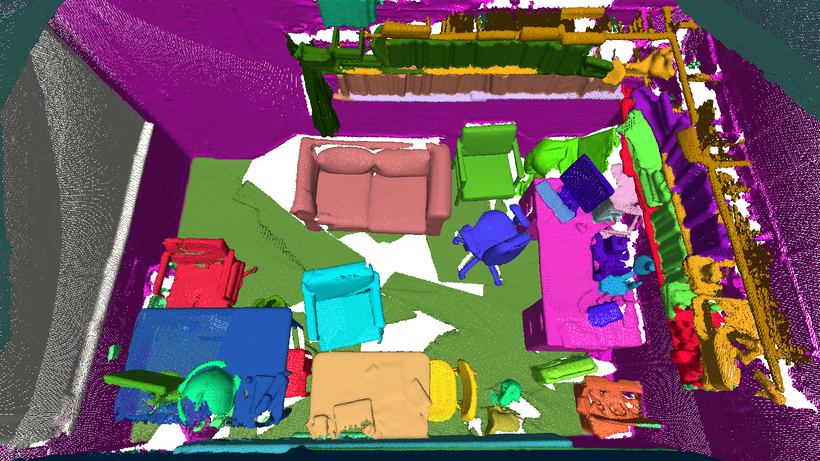}
    \centerline{Ground truth}
  \end{minipage}
  \hspace{\hsp}
  \begin{minipage}{0.23\linewidth}
    \includegraphics[width=\linewidth]{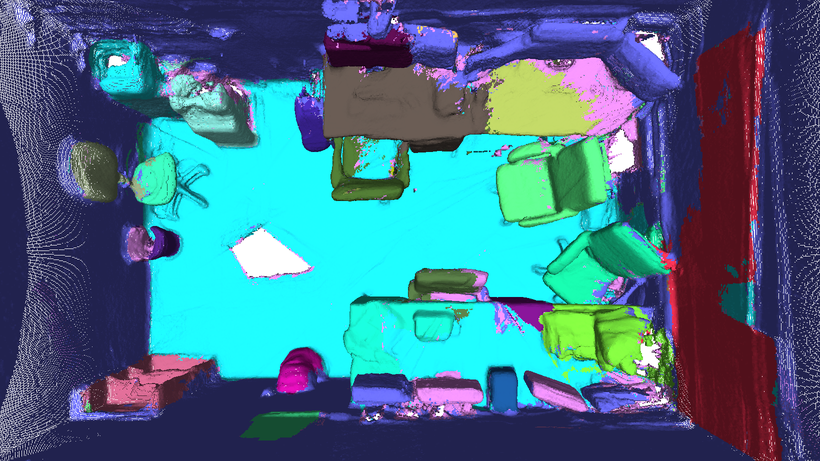}
    \centerline{Prediction}
  \end{minipage}%
  \begin{minipage}{0.23\linewidth}
    \includegraphics[width=\linewidth]{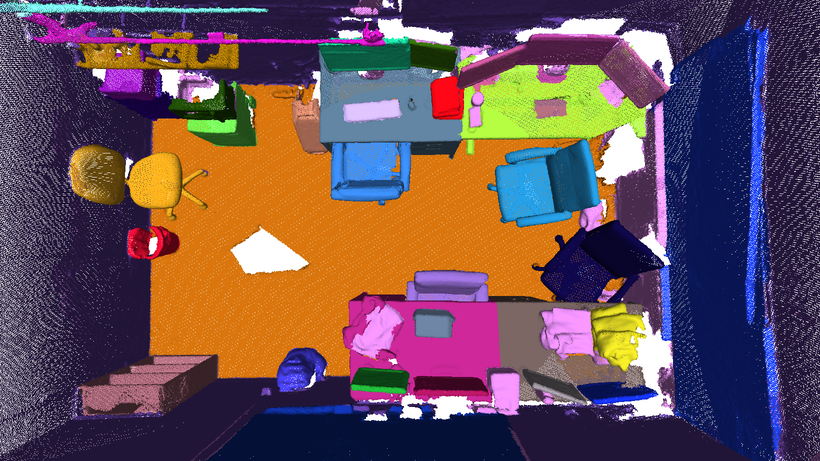}
    \centerline{Ground truth}
  \end{minipage}%
  \caption{Instance-based semantic segmentation on SceneNN
    dataset~\cite{hua-scenenn-3dv16}.}
  \label{fig:instance}
  \vspace{-0.2in}
\end{figure*}

\begin{figure}[ht]
  \centering
  \includegraphics[width=0.8\linewidth]{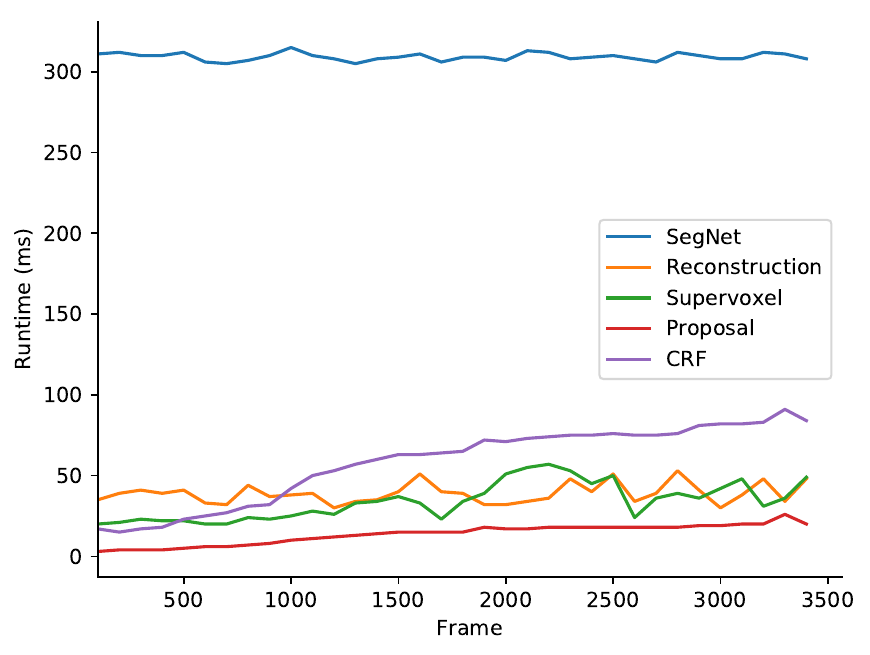}
  \caption{Runtime analysis of our progressive semantic segmentation system.}
  \label{fig:runtime}
  \vspace{-0.2in}
\end{figure}

\vspace{-0.1in}
\paragraph{Runtime analysis.}
Runtime analysis is performed on a desktop with an Intel Core i7-5820K 3.30GHz
CPU, 32GB RAM, and an NVIDIA Titan X GPU. The average runtime breakdown of each
step in the pipeline is demonstrated in Figure~\ref{fig:runtime}.  Specifically,
it takes $309.3$ms on average to run a single forward pass of neural
network. Building super-voxels takes $34.1$ms. CRF with higher-order constraints
requires additional $57.9$ ms. As can be seen, over time when more regions in
the scene are reconstructed, our semantic segmentation still takes constant
running time on average.

We compared our online approach to the reference offline approach that runs CNN
prediction every frame (Table~\ref{tab:offline}). We see that the accuracy of
our online method (Table~\ref{tab:online}) is about 5\% lower on average, but
the speed gain is more than 8 times. Our system runs at 10-15Hz. With the same
CNN predictions, direct fusion method~\cite{cavallari-semanticfusion-eccvw16}
runs at 17-20Hz, and SemanticFusion~\cite{mccormac-semanticfusion-icra17} runs
at 14-16Hz. Note that such methods do not constrain label consistency.

\vspace{-0.2in}
\paragraph{Temporal accuracy.}
Since our method can be run in real time, we evaluate the segmentation accuracy
over time. For every scene, we measure the accuracy every $100$ frames. The
progressive segmentation results are shown in Figure~\ref{fig:seg3d}.

The results suggest that our method \emph{consistently} outperforms other
methods in a long run, not just only at a certain time period. In addition, we
observe that the accuracy over time sometimes still fluctuates slightly due to
the lack of full temporal constraints among the CNN predictions. Addressing this
issue could be an interesting future work.

\begin{figure*}[t!]
  \begin{minipage}{\linewidth}
    \center
    \includegraphics[width=0.9\linewidth]{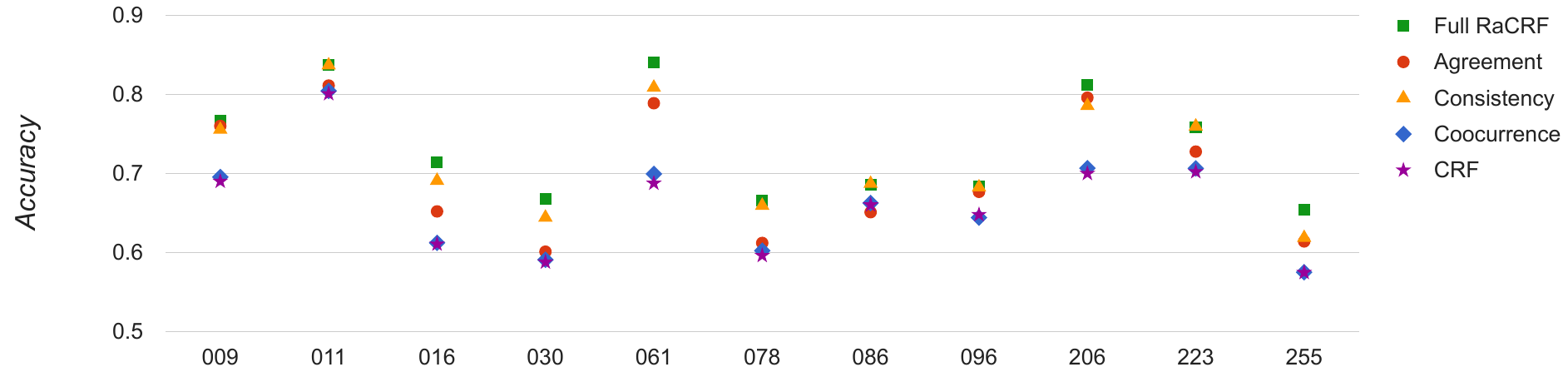}
  \end{minipage}
  \caption{Ablation study on the effects of different CRF terms. There is
    usually a noticeable gap between the performances of the conventional dense
    CRF and ours. In addition, individual term helps improving the segmentation
    accuracy. This study also shows the importance of consistency in semantic
    segmentation.}
  \label{fig:ablation}
\end{figure*}

\vspace{-0.1in}
\paragraph{Ablation study.}
To further understand the performance of our CRF model, we carry out an ablation
study to evaluate the effects of each CRF term on the result segmentation.  We
execute three runs on $10$ scenes, each run enables only one term in our CRF
model, and record their performances. Figure \ref{fig:ablation} visualizes the
results on these $10$ scenes. In general, running full higher-order model
achieves the best performance. Enabling individual term is able to outperform
the base dense CRF model. The consistency term contributes the most in the
performance boost, which validates our initial hypothesis that object-level
information is crucial when performing dense semantic segmentation.

\vspace{-0.1in}
\paragraph{Instance segmentation.}
To evaluate our instance segmentation results, We use the average precision
metric \cite{lin-coco-eccv14} with minimal 50\% overlap. The results are shown
in Table \ref{tab:online}. Figure \ref{fig:instance} visualizes the instance
segmentation in two indoor scenes using our approach. Such results could serve
as a baseline to compare with more sophisticated real-time 3D instance
segmentation technique in the future.

%% file: remark.tex
\section{Conclusion}
\vspace{-0.1in} Our proposed system demonstrates the capability to integrate
semantic segmentation into real-time indoor scanning by optimizing the
predictions from a 2D neural network with a novel higher-order CRF model. The
results and ground truth category-based and instance-based semantic segmentation
will be made publicly available. The results from our system can further be used
in other interactive or real-time applications, e.g., furniture arrangement
\cite{yu-furniture-sg11}, or object manipulation and picking in robotics.

\vspace{-0.1in}
\paragraph{Acknowledgment.} This research project is partially supported by an
internal grant from HKUST (R9429).